\documentclass{bmvc2k}

\usepackage{subfig}
\usepackage{graphicx,wrapfig,lipsum}

\usepackage{graphicx}
\usepackage{amsmath}
\usepackage{amssymb}
\usepackage{booktabs}

\usepackage{bbold}
\usepackage{bm}
\usepackage{pgfplots}
\usepackage{enumerate}
\usepackage{enumitem}
\usepackage{caption}
\usepackage{colortbl}
\usepackage{multirow}
\usepackage{pgf-pie}
\usepackage{xcolor}
\usetikzlibrary{shadows}
\usepackage{mathtools}
\usepackage{booktabs}
\usepackage{amssymb}
\usepackage{amsmath}
\usepackage{float}
\usepackage{soul}
\usepackage{pifont} 

\title{Syn-to-Real Unsupervised Domain Adaptation for Indoor 3D Object Detection}

\addauthor{Yunsong Wang}{yunsong@comp.nus.edu.sg}{1}
\addauthor{Na Zhao}{na_zhao@sutd.edu.sg}{2}
\addauthor{Gim Hee Lee}{gimhee.lee@nus.edu.sg}{1}

\addinstitution{
\footnotesize
 School of Computing\\
 National University of Singapore\\
 Singapore
}
\addinstitution{
\footnotesize
 Information Systems Technology and Design\\
 Singapore University of Technology and Design\\
 Singapore
}

\runninghead{AUTHORS}{YUNSONG WANG, NA ZHAO, GIM HEE LEE}

\begin{document}

\maketitle

\begin{abstract}
The use of synthetic data in indoor 3D object detection offers the potential of greatly reducing the manual labor involved in 3D annotations and training effective zero-shot detectors.
    However, the complicated domain shifts across syn-to-real indoor datasets remains underexplored. 
    In this paper, we propose a novel Object-wise Hierarchical Domain Alignment (OHDA) framework for syn-to-real unsupervised domain adaptation in indoor 3D object detection. Our approach includes an object-aware augmentation strategy to effectively diversify the source domain data, and we introduce a two-branch adaptation framework consisting of an adversarial training branch and a pseudo labeling branch, in order to simultaneously reach holistic-level and class-level domain alignment.
    The pseudo labeling is further refined through two proposed schemes specifically designed for indoor UDA. 
        Our adaptation results from synthetic dataset 3D-FRONT to real-world datasets ScanNetV2 and SUN RGB-D demonstrate remarkable mAP$_{25}$ improvements of $9.7\%$ and $9.1\%$ over Source-Only baselines, respectively, and consistently outperform the methods adapted from 2D and 3D outdoor scenarios. The code is available at \href{https://github.com/wangys16/OHDA}{\textbf{https://github.com/wangys16/OHDA}}.
\end{abstract}

\section{Introduction}
\label{sec:intro}

\begin{wrapfigure}{r}{7cm}
  \centering
  \raisebox{0pt}[\dimexpr\height-0.6\baselineskip\relax]{\includegraphics[width=1\linewidth]{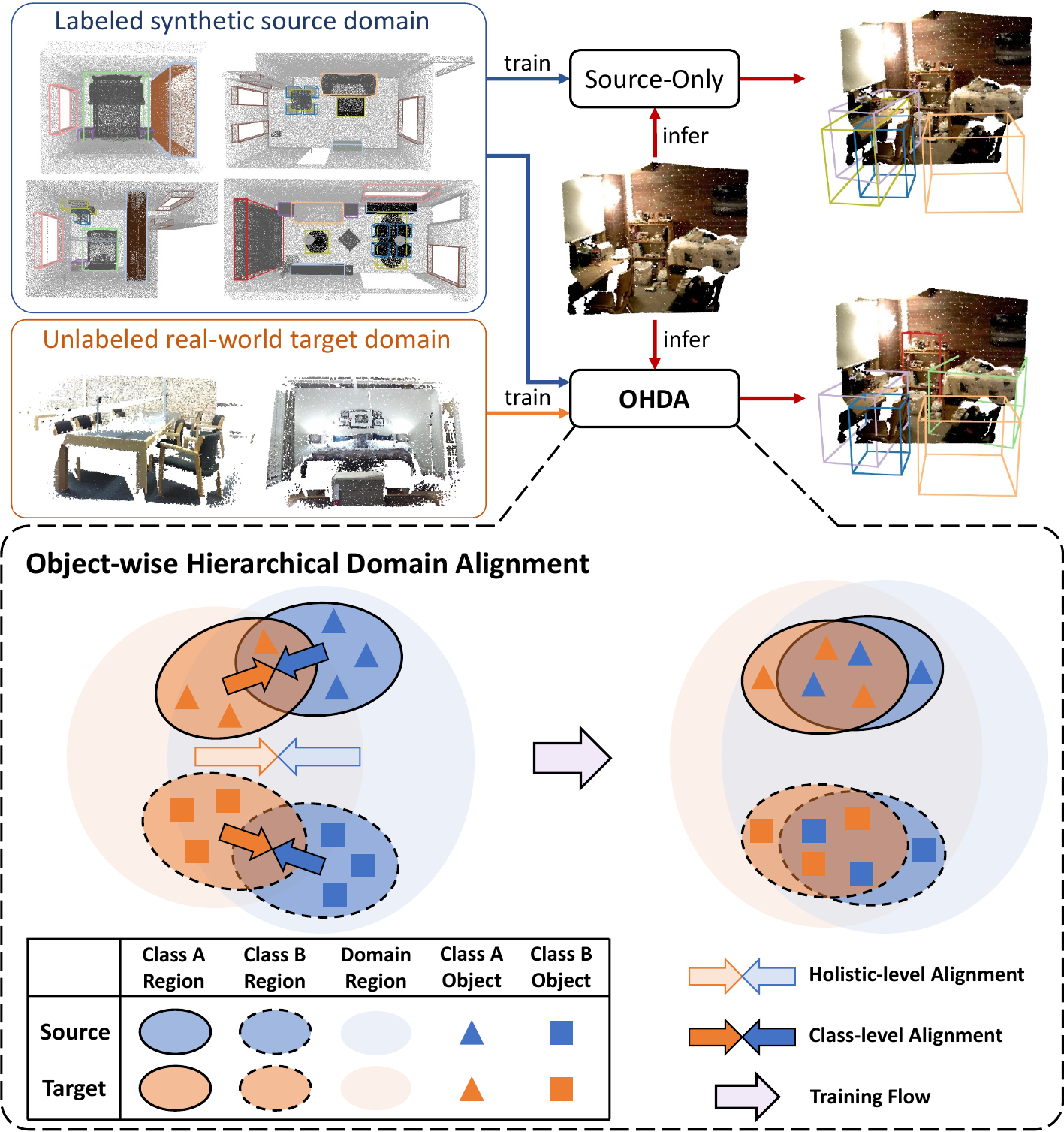}}
  \vspace{-0.15in}
  \caption{\textbf{Pipelines of Source-Only and OHDA.} 
    }
    \label{fig:teaser}
  \label{curves}
\end{wrapfigure}

3D object detection is a fundamental yet challenging task that plays a crucial role in numerous applications such as autonomous driving and augmented reality. 
In recent years, several deep learning approaches \cite{3d-sis,votenet,h3dnet,group-free,hyperdet} developed for 3D object detection have achieved promising performance. 
However, the performance heavily relies on 
large amounts of well-annotated training data, 
and an in-distribution assumption between training and testing data. When the testing environment (target domain) follows a different distribution from the training environment (source domain), the performance would suffer from a significant drop, as shown by the prediction from the source-only model in Figure \ref{fig:teaser}. This out-of-distribution phenomenon is caused by the domain shift 
due to a variety of reasons such as different sensor configurations \cite{tsai2022see} or geographical locations \cite{germany} during training and testing data collection.

To deal with the domain shift between the training and testing environments, unsupervised domain adaption (UDA) is proposed by learning domain-invariant knowledge from both the labeled source domain data and the unlabeled target domain data. Recently, a few domain adaptive 3D object detection methods \cite{sf-uda,germany,st3d,adversarial} have been devised to avoid requiring heavy 3D annotations on the target domains. Nonetheless, these works predominantly focus on outdoor 3D object detection. 
Directly applying these techniques to indoor settings, as referenced in Table \ref{scannet} and Table \ref{sunrgbd}, often yields suboptimal outcomes.
The root of such underperformance mainly lies in 
the distinct domain shifts characterizing outdoor and indoor contexts.
Outdoor datasets normally focus on detecting cars and the major domain shift is the average sizes of cars \cite{germany}. In contrast, indoor environments present intricate domain gaps across diverse object categories, 
and demand further investigations.

To this end, we focus on unsupervised domain adaptation for indoor 3D object detection. 
Compared to outdoor scenarios, different indoor datasets can have much larger and more complicated cross-domain gaps due to different capturing patterns (single-view \textit{vs} multi-view) and diverse object categories with various shape distributions, which lead to the difficulty of using real-world datasets as source domain. In contrary, synthetic data offers large quantity of complete scene-level point clouds with labor-free 3D labels, covering common object categories with diverse shapes, which is suitable for the challenge of indoor UDA and has the potential of training an effective zero-shot 3D detectors without real-world 3D labels. Therefore, in this work we study the Synthetic-to-Real Unsupervised Domain Adaptation (SR-UDA) for indoor 3D object detection.

We adopt the synthetic datatset 3D-FRONT \cite{3d-front} as the source domain, which contains 18,968 rooms furnished by 13,151 furniture objects with high-quality textures.
To narrow down the syn-to-real domain gap, we introduce an object-aware augmentation strategy, which effectively diversifies the synthetic scenes and is seamlessly integrated with a virtual scan simulation (VSS) technique \cite{doda} for data augmentation. 
To adapt the knowledge across syn-to-real domain shift, we propose an Object-wise Hierarchical Domain Alignment (OHDA) framework, which comprises of an proposal-level domain adversarial training branch and a pseudo-labeling branch, which aligns the cross-domain proposal features and performs more effective joint training on cross-domain data. We further refine the pseudo labeling through two proposed modules: a simple-yet-effective adaptive thresholding strategy to resist the class-specific domain gaps, and a reweighting scheme on pseudo labels based to their consistency towards model perturbation. 
We assess our methodology using two benchmark datasets for indoor 3D object detection, namely ScanNetV2 \cite{scannet} and SUN RGB-D \cite{sunrgbd}. In these evaluations, our OHDA method consistently showcases significant improvements. In summary, our key \textbf{contributions} include:
\begin{itemize}
[leftmargin=2ex, topsep=1pt, itemsep=-1ex]
    \item 
    To the best of our knowledge, 
    this is the first study that delves into UDA for indoor 3D object detection using a syn-to-real approach, which eliminates the need for large-scale 3D labels that are notably costly and challenging to acquire for 3D object detection.
    \item We propose OHDA as the first solution to the SR-UDA task, which benefits from the two-branch structure that leads to holistic-level and class-level domain alignment.
    We further propose specific strategies to adapt OHDA to the SR-UDA setting, including object-aware augmentation and two pseudo label refinement modules.
    \item 
    We establish new benchmarks for 3D-FRONT $\rightarrow$ ScanNetV2 and 3D-FRONT $\rightarrow$ SUN RGB-D, improving mAP$_{25}$ by $9.7\%$ and $9.1\%$ over the Source-Only baseline, respectively, consistently surpassing existing UDA methods adapted from different contexts.
\end{itemize}
\section{Related Work}

\noindent\textbf{Indoor 3D Object Detection.} Benefiting from the well preserved spatial information in point clouds, methods that directly process on point clouds have become the majority in state-of-the-art indoor 3D object detection. Early methods transform the raw 3D point clouds into voxels \cite{voxnet,voxelnet} or use the 2D priors \cite{frustum,2d-driven} to tackle the difficulties of working on 3D data. Inspired by Hough Voting in 2D object detection \cite{hough1,hough2}, VoteNet \cite{votenet} proposes to predict the offsets from sampled points to their corresponding bounding box centers, which is followed by clustering and PointNet \cite{pointnet++}-based grouping. Following VoteNet, several recent methods further incorporate 3D primitives \cite{h3dnet}, back-tracing strategy \cite{brnet}, or object-level attention \cite{group-free}. 

\vspace{2pt}
\noindent\textbf{Unsupervised Domain Adaptation.} Given the labeled source domain data and the unlabeled target domain data, unsupervised domain adaptation (UDA) aims to
effectively adapt knowledge from the source to target domain.
This task has been largely explored in 2D, where the discrepancy-based methods \cite{discrep1,discrep2,discrep3,discrep4} learn to minimize the domain discrepancy, the domain adversarial training methods \cite{adv1,adv2,adv3,adv4,adv5} address it by jointly training a domain classifier with min-max optimization, and self-training (\textit{i.e.} pseudo labeling) methods \cite{st1,st2,st3,st4,st5} iteratively train the model using both the labeled source and pseudo-labeled target domain data. Most existing methods focus on unsupervised domain adaptation for image classification and semantic segmentation tasks, which are less challenging than the object detection task that requires the precise regression of the bounding boxes.

\vspace{2pt}
\noindent\textbf{Domain Adaptive Object Detection.} For domain adaptive 2D object detection, most of previous work leverage domain adversarial training to align the cross-domain global feature or proposals \cite{obj_adv1,obj_adv2,obj_adv3,obj_adv4}. Recently, several approaches also convert this task to the semi-supervised learning problem and utilize advanced pseudo labeling techniques such as Mean Teacher \cite{mt1,mt2,wang20213dioumatch} to address the domain shift problem. For domain adaptive 3D object detection, existing works mainly address it under the outdoor scenarios, through bounding box size adaptation \cite{germany}, pseudo label time-consistency \cite{sf-uda}, Mean Teacher \cite{st3d} and global-level domain adversarial training \cite{adversarial}. In contrast, the UDA of 3D object detection in indoor scenarios, which consist of more diverse object categories with class-specific domain gaps across datasets, is more challenging and pragmatic yet underexplored. Thus, this work focuses on this pragmatic task and adopts a large-scale and manual-annotation-free synthetic dataset as the source domain.  


\begin{figure*}[htbp]
    \centering
    \includegraphics[width=1\textwidth]{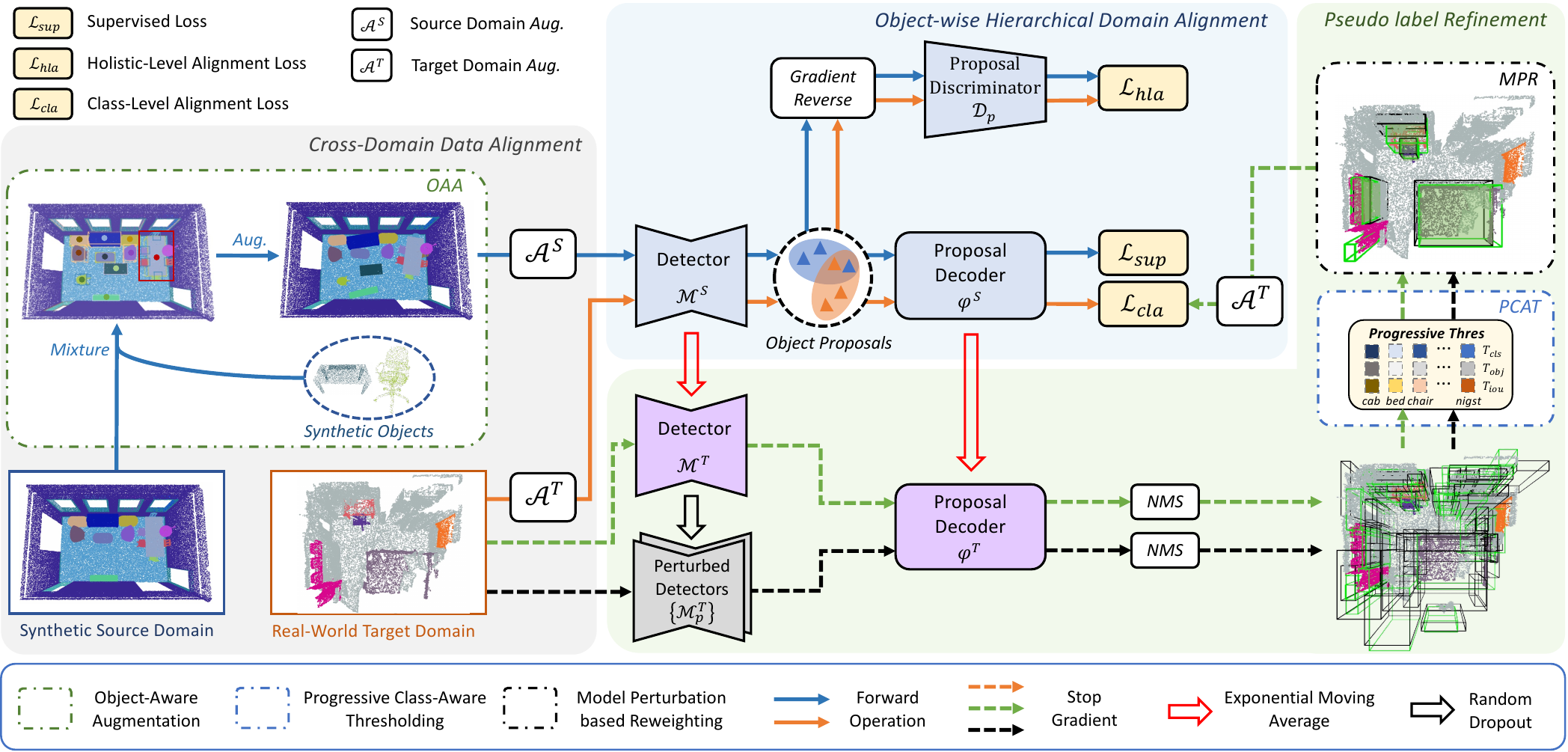}
    \vspace{-0.15in}
    \caption{\textbf{Proposed Framework.} 
    Our framework consists of three parts: cross-domain data alignment, Object-wise Hierarchical Domain Alignment (consisting of Class-Level Alignment (CLA) and Holistic-Level Alignment (HLA)), and Pseudo Label Refinement (PLR).
    }
    \label{fig:framework}
\end{figure*}

\section{Our Approach}

\subsection{Overview}
The framework of OHDA is shown in Figure \ref{fig:framework}. 
Formally, we denote the detector backbone as $\bm{\mathcal{M}}$, which predicts the object proposals, and $\bm{\varphi}$ is the proposal decoder that predicts the bounding box attributes. We denote the student and teacher model as $\bm{\varphi}^S(\bm{\mathcal{M}}^S(\cdot))$ and $\bm{\varphi}^T(\bm{\mathcal{M}}^T(\cdot))$, respectively. 
Both student and teacher are initialized by a pre-trained model trained on the augmented source domain data. During adaptation, the student model is updated using gradient decent, and the teacher model is detached and is updated by Exponential Moving Average (EMA) of the student model. 
In SR-UDA, we are given $N_s$ labeled point clouds from the synthetic source domain $\bm{\mathcal{D}}_s=\{\bm{X}_s, \bm{Y}_s\}^{N_s}$ and $N_t$ unlabeled point clouds from the real-world target domain $\bm{\mathcal{D}}_t=\{\bm{X}_t\}^{N_t}$.
\subsection{Cross-Domain Data Alignment}\label{sec:OAA}

The syn-to-real domain gaps in indoor 3D scenes are mainly caused by the change of two aspects:
the RGB-D camera scanning pattern and the room layout. The first 
type of domain gap can be largely solved by VSS \cite{doda}, which is a recent technique that simulates sensor noise and occlusion similarly to real-world scenes. In this paper, we propose an \textit{Object-Aware Augmentation} (OAA) strategy that can narrow down the second type of domain gap and is compatible with VSS. 

\vspace{2pt}
\noindent\textbf{Scene-Object Mixture.} Although 3D-FRONT benefits from its large quantity of individual scenes, it still suffers from \textit{regular localization} of objects. Besides, the \textit{class-imbalance} problem can also lead to inferior performance of detector. We therefore propose to mix the scenes and the synthetic objects. Specifically, we randomly sample objects from 3D-FRONT \cite{3d-front} and randomly place them into the scene on-the-fly. The objects are placed on the floor, with no collision with original scene. Furthermore, we set the probability of sampling each object category to be inversely proportional to their frequencies in 3D-FRONT \cite{3d-front}, so as to mitigate the class-imbalance problem.

\vspace{2pt}

\vspace{2pt}
\noindent\textbf{Local Pattern Preserved Augmentation.}
To diversify the synthetic source domain, the local augmentation on object level is essential, which includes random scaling, rotating and translating the objects on the floor. To prevent the generation of unrealistic scenes, it is crucial to perform collision checks between objects during object augmentation. However, collision checks in indoor scenes are more complex than those in outdoor scenes \cite{localaug} due to the intricate local spatial patterns between objects. For example, some objects such as chairs and tables have naturally overlapping bounding boxes in the original scenes. Therefore, directly checking the overlap of bounding boxes may not be sufficient to validate object collision. To address this challenge, we propose preserving local patterns by merging overlapped objects. Specifically, before augmentation, we identify existing collided bounding boxes (where IoU$>0.01$) and merge all points from the corresponding collided bounding boxes to obtain new object points. The collided bounding boxes are replaced by a minimum axis-aligned bounding box that covers them. We then iteratively merge the collided objects until there are no more collisions, followed by local augmentation on the merged objects.



\subsection{Object-wise Hierarchical Domain Alignment}
To enhance the domain adaptation under the indoor scenario with more complicated object distributions and cross-domain shifts, we propose a hierarchical domain alignment framework performed on object-level, which includes class-level and holistic-level alignment.
\subsubsection{Class-Level Alignment}

We first build a pseudo labeling branch to facilitate the domain alignment using labeled source domain data and unlabeled target domain data. 
Formally, the supervised loss on the labeled source domain samples is computed as $\mathcal{L}_{sup} = \mathcal{L}^{l}\left(\hat{\bm{Y}}_{s}, \bm{Y}_s\right)$, where $\hat{\bm{Y}}_{s}=\bm{\varphi}^S\left(\bm{\mathcal{M}}^S(\bm{X}_s)\right)$ are the student predictions on source domain samples. The Class-Level Alignment (CLA) loss on the pseudo-labeled target domain data is computed as:  
\vspace{-1mm}
\begin{gather}
\label{cla}
\mathcal{L}_{cla} = \frac{1}{M}\sum_{m=1}^M\mathcal{L}^u\left(\hat{\bm{Y}}_t^m, \bm{\Bar{Y}}_t^m\right),\\
\bm{\hat{Y}}_t^m=\left[\bm{\varphi}^S(\bm{\mathcal{M}}^S(\bm{X}_t))\right]^m, 
\Bar{\bm{Y}}_t^m=\mathcal{S}\left(\mathcal{R}(\bm{\varphi}^T(\bm{\mathcal{M}}^T(\bm{X}_t))), \bm{\hat{Y}}_t^m\right),
\vspace{-2mm}
\end{gather}
where $M$ the number of proposals, $\mathcal{F}(\cdot)$ the pseudo label refinement strategy (as discussed in Sec \ref{sec:plr}), $\bm{\hat{Y}}_t^m$ the $m$-th bounding box predicted by student model, $\bm{\Bar{Y}}_t^m$ the teacher's prediction that is matched with $\bm{\hat{Y}}_t^m$ using distance-based scheme $\mathcal{S}$ similarly to \cite{votenet}. We supervise the student's predictions that are within $0.3m$ of pseudo bounding box centers. 
$\mathcal{L}^{l}$ is the standard supervised loss used in \cite{votenet}, and $\mathcal{L}^u$ only contains the box loss and the semantic loss.

\subsubsection{Holistic-Level Alignment}

The aforementioned Class-Level Alignment is similar to the standard mean teacher \cite{mt} in 2D semi-supervised learning, which however could lead to suboptimal performance due to the domain gap between synthetic and real domains under UDA setting. Therefore, we propose to further encourage the cross-domain proposals to share the same feature space, such that the ground truth labels on the source domain and pseudo labels on the target domain 
can achieve better joint-supervision on the student model. 
Specifically, we add a Proposal Discriminator $\bm{\mathcal{D}_p}(\cdot)$ to predict the domain label (0 for source, 1 for target domain) on the object proposal level. The Holistic Level Alignment (HLA) loss is implemented as a cross-entropy loss on all object proposals:  
\begin{equation}
    \mathcal{L}_{hla}=
    \max_{\bm{\mathcal{M}}^S}\min_{\bm{\mathcal{D}_p}}\left\{-\text{log}\left(1-\bm{D}^S\right)-\text{log}\left(\bm{D}^T\right)\right\},\\
\label{hla}
\end{equation}
%
%
where $\bm{D}^S=\bm{\mathcal{D}_p}(\bm{\mathcal{M}}^S\left(\bm{X}_s\right))$ and $\bm{D}^T=\bm{\mathcal{D}_p}(\bm{\mathcal{M}}^S(\bm{X}_t))$ are the predicted possibilities of the aggregated proposal features $\bm{\mathcal{M}}^S\left(\bm{X}_s\right), \bm{\mathcal{M}}^S(\bm{X}_t)$ coming from target domain.


\subsection{Pseudo Label Refinement}
\label{sec:plr}
\subsubsection{Progressive Class-Aware Thresholding}
\label{sec:PCAT}
One issue that arises when devising the mean teacher paradigm under UDA settings for indoor 3D object detection is the occurrence of class-specific domain gaps, 
for which we first develop a straightforward \textit{class-aware thresholding} strategy that accounts for various confidence distributions during the pseudo label filtering to handle the class-specific domain gaps. Initially, all target domain data is fed into the initialized teacher model to predict pseudo labels, and we store the per-class pseudo label confidence scores:
\begin{equation}
    \left[\bm{Q}_{obj}^c, \bm{Q}_{cls}^c, \bm{Q}_{iou}^c\right]=\left[\bm{\tau}_{obj}^c(\bm{\Bar{Y}}_t), \bm{\tau}_{cls}^c(\bm{\Bar{Y}}_t), \bm{\tau}_{iou}^c(\bm{\Bar{Y}}_t)\right],
\end{equation}
where $\bm{\tau}_{obj}^c(\bm{\Bar{Y}}_t)$, $\bm{\tau}_{cls}^c(\bm{\Bar{Y}}_t)$, $\bm{\tau}_{iou}^c(\bm{\Bar{Y}}_t)$ denote the objectness score, maximum classification score, and IoU score of the proposals in $\bm{\Bar{Y}}_t$ with class $c$, respectively. We calculate the class-aware thresholds as:
\begin{equation}
    \label{cat}
    \bm{T}_{c, m}=\operatorname{Clamp}\left(\operatorname{Percentile}_{\alpha}\left(\bm{Q}_{m}^c\right), T_{l}^m, T_{h}^m\right),
\end{equation}
where $m\in\{obj, cls, iou\}$ is the confidence metric, $\operatorname{Percentile}_{\alpha}\left(\cdot\right)$ is percentile function that outputs the value higher than $\alpha\%$ of input, and $\operatorname{Clamp}(\cdot,T_{l}, T_{h})$ is clamping between $T_{l}^m$ and $T_{h}^m$. 
On top of the class-aware thresholds, we further perform progressive updates 
such that the thresholds are dynamically fit to the pseudo label confidence distributions, which leads to our \textit{Progressive Class-Aware Thresholding (PCAT)}. 
Denoting $\bm{\Bar{T}}_{c, m}^k$ as the progressive class-aware threshold for epoch $k$, 
we update it by:
\begin{equation}
    \bm{\Bar{T}}_{c, m}^k = \begin{cases}
        \quad\quad\quad\quad\bm{T}_{c, m}, &k=0, \\
        \beta\cdot\bm{\Bar{T}}_{c, m}^{k-1}+(1-\beta)\cdot\bm{T}_{c, m}^{k-1}, &\text{otherwise}.
    \end{cases}
\end{equation}
%
Here, $T_{c, m}^{k-1}$ is computed with Eq.~\eqref{cat} using the confidence scores in epoch $k-1$, and
$\beta$ is the update momentum. 




\subsubsection{Model Perturbation based Reweighting}

Although the aforementioned PCAT enables adaptively filtering pseudo labels for diverse object classes, the pseudo labeling loss still assigns uniform weights for all valid pseudo labels after score filtering. To further mine the high-quality pseudo labels, we propose the Model Perturbation based Reweighting (MPR) module, which reweights the pseudo labels according to the pseudo label consistency.
Specifically, given each batch of data, we first perform random dropout on the teacher detector backbone to get a set of perturbed teacher models $\{\bm{\mathcal{M}}^T_{p}\}_{p=1}^{P}$, where $P$ is the number of perturbed models. We then feed it with target domain data and go through Non-Maximum Suppression (NMS) and PCAT similarly to get the perturbed pseudo labels $\{\Bar{\bm{Y}}_{t,p}\}_{p=1}^{P}$. Consequently, we assess the robustness of $\Bar{\bm{Y}}_t$ according to their IoU with $\{\Bar{\bm{Y}}_{t,p}\}_{p=1}^{P}$, such that each pseudo bounding box is reweighted based on their consistency \textit{w.r.t} model perturbation. Formally, we compute:

\vspace{-3mm}
\begin{gather}
\label{cla}
\bm{\mathcal{U}}^m=\frac{1}{P}\sum_{p=1}^P\max_{n}\ \text{IoU}(\Bar{\bm{Y}}_t^m, \Bar{\bm{Y}}_{t,p}^n),\; \bm{\omega}(\bm{\mathcal{U}}^m)=1+\lambda_{mpr}\cdot\bm{\mathcal{U}}^m,\\
\begin{split}
    \mathcal{L}_{cla}^{mpr} = &\frac{\sum_{m=1}^M\bm{\omega}(\bm{\mathcal{U}}^m)\cdot\mathcal{L}^u\left(\hat{\bm{Y}}_t^m,
    \bm{\Bar{Y}}_t^m\right)}{\sum_{m=1}^M\bm{\omega}(\bm{\mathcal{U}}^m)}.
\end{split}
\end{gather}
Consequently, the MPR is aimed at mining the consistent pseudo labels, which is seamlessly integrated with the score filtering to further enhance pseudo labeling.

\subsection{Training Objective}

The overall loss is given by:

\vspace{-1mm}
\begin{equation}
    \mathcal{L} = \mathcal{L}_{sup}+\lambda_{hla}\mathcal{L}_{hla}+\lambda_{cla}\mathcal{L}_{cla}^{mpr},
\end{equation}
where $\lambda_{hla}$ and $\lambda_{cla}$ are weights to balance different losses. The losses work synergically on object proposal level, aiming to achieve hierarchical cross-domain alignments.

\section{Experiments}

\subsection{Implementations Details}
\label{detect_detail}

\noindent\textbf{Dataset settings.} For the data preprocessing of 3D-FRONT \cite{3d-front}, we obtain point clouds by uniformly sampling on the mesh surfaces, and select scenes with $3\sim 20$ interested objects, getting 10,515 scenes for training and 1,104 scenes for validation. As for the label mapping, we select 10 and 7 object categories for 3D-FRONT $\rightarrow$ ScanNetV2 and 3D-FRONT $\rightarrow$ SUN RGB-D settings, respectively.

\vspace{2pt}
\noindent\textbf{Experiment settings.} 
We leverage VoteNet \cite{votenet} as the detector and first train it on 3D-FRONT for 50 epochs, 
then adapt it to target domains ScanNetV2 and SUN RGB-D individually. 
We regard traversing target domain as one epoch, and respectively train 100 and 50 epochs for ScanNetV2 and SUN RGB-D.
\subsection{Baselines}
\label{baselines}

We set up baselines under source-only and UDA settings. 
For UDA setting, we use OAA+VSS for source domain augmentation. Since there is no prior work focusing on UDA for indoor 3D object detection, we set up the baselines adapted from 2D and outdoor detection:

\noindent\textbf{Naive MT \cite{mt}}: Based on the mean teacher paradigm, we reproduce the losses in VoteNet \cite{votenet} for the unlabeled target domain data, using the pseudo labels provided by the mean teacher.
\begin{table*}[t]
    \footnotesize
    \centering
    \setlength{\tabcolsep}{1.35mm}{
    \begin{tabular}{c|c|cccccccccc|cc}
        \toprule

         &Method&cab& bed& chair& sofa& tabl& door& wind& bkshf& desk & nigst& mAP$_{25}$ & mAP$_{50}$ \\
         \midrule[0.8pt]
         \midrule
         \parbox[t]{2mm}{\multirow{3}{*}{\rotatebox[origin=c]{90}{SO}}}&\emph{w/o Aug.} \cite{votenet}&2.1 & 50.1& 63.9& 45.7& 36.3& \underline{2.5}& 0.7& 25.1& 23.1& 15.8 & 26.5 & 13.5\\

         &VSS \cite{doda}&9.3 & \underline{61.3}& 55.2& 67.3 & 37.4& 2.4& 4.8& 28.3& 25.7& 40.6 & 33.2 & 19.1\\
         &OAA+VSS&9.7 & \textbf{61.7}& 55.3 & 74.8 & 41.0& 1.8& 4.4& 29.9& 29.8& 42.6 & 35.1 & 20.3\\
         \midrule
         \parbox[t]{2mm}{\multirow{4}{*}{\rotatebox[origin=c]{90}{UDA}}}
         &Naive MT$^\dag$ \cite{mt}&7.6 & 60.7& 65.5& 63.3 & 40.4& 2.1& 3.3& 30.5& 11.9& 44.3 & 33.0 & 20.0\\
         &ST3D$^\dag$ \cite{st3d}&10.7 & 54.9& 66.5& 77.0 & 39.2& 3.0& 5.9& \underline{32.7}& 27.1& 49.1 & 36.3 & 22.6\\
         &Global-DAT$^\dag$ \cite{adversarial}&6.0 & 63.4& 57.3& 66.6 & \underline{42.9}& 1.6& 3.6& 22.3& \underline{40.7}& 25.3 & 32.9 & 19.3\\
         &Ours \textit{w/o} PLR& \underline{12.0} & 60.8 & \underline{69.1} & \underline{77.9} & 41.3 & 2.0 & \underline{6.5} & 31.2 & 29.2 & \underline{55.2} & \underline{38.5} & \underline{23.9} \\
         &\textbf{Ours}&\textbf{12.2} & 58.8 & \textbf{71.0}& \textbf{80.2} & \textbf{45.7}& \textbf{4.4}& \textbf{16.0}& \textbf{35.4}& \textbf{42.7}& \textbf{64.6} & \textbf{42.9} & \textbf{27.6}\\

         \midrule
         \parbox[t]{2mm}{\multirow{1}{*}{\rotatebox[origin=c]{90}{FS}}}&Oracle&32.0 & 87.4& 86.3& 88.3 & 55.5& 39.1& 34.7& 39.0& 69.6& 76.7 & 60.9 & 39.9\\
        
         \bottomrule
    \end{tabular}
    }
    \captionsetup{font=small}
    \vspace{-0.01in}
    \caption{\small \textbf{3D-FRONT $\rightarrow$ ScannetV2 with per object category mAP@0.25 results.} We compare the results under Source-Only (SO), UDA and Fully-Supervised (FS) settings. \textit{Aug.} denotes data augmentations. 
    We indicate the best and runner-up adaptation results by \textbf{bold} and \underline{underline}, respectively.}
    \label{scannet}
    \vspace{-1mm}
\end{table*}
\begin{table*}[htbp]
    \footnotesize
    \centering
    \setlength{\tabcolsep}{2.4mm}
    {
    \begin{tabular}{c|c|ccccccc|cc}
        \toprule

         &Method&bed& tabl& sofa& chair& desk& nigst& bkshf & mAP$_{25}$ & mAP$_{50}$ \\
         \midrule[0.8pt]
         \midrule
         \parbox[t]{2mm}{\multirow{3}{*}{\rotatebox[origin=c]{90}{SO}}}&\emph{w/o Aug.} \cite{votenet}&35.3 & 28.1& 19.6& 40.2& 7.3& 1.4& 4.4& 19.5 & 6.4\\
         &VSS \cite{doda}& 58.2& 30.9& 39.9 & 45.5& 11.7& 5.4& 5.4& 28.2 & 11.0\\
         &OAA+VSS&61.5 & 29.9& 42.4& 45.2 & \textbf{13.8}& 15.2& 4.8& 30.4 & 14.1\\
         \midrule

         \parbox[t]{2mm}{\multirow{4}{*}{\rotatebox[origin=c]{90}{UDA}}}
         &Naive MT$^\dag$ \cite{mt}& 64.4 & 27.3 & 37.9 & 41.0 & 5.3 & 26.2 & 4.1 & 29.5 & 10.6 \\
         &ST3D$^\dag$ \cite{st3d}& 70.3 & 33.4 & 37.7 & 50.4 & 11.4 & 26.9 & 5.8 & 33.8 & 15.1 \\
         &Global-DAT$^\dag$ \cite{adversarial}& 55.3 & 27.9 & 37.6 & 39.8 & 8.9 & 5.1 & 4.9 & 25.6 & 9.6 \\
         &Ours \textit{w/o} PLR& \underline{71.8} & \underline{35.7} & 42.5 & \underline{54.0} & 8.5 & \textbf{29.4} & \underline{6.2} & \underline{35.5} & \underline{15.6} \\
         &\textbf{Ours}&\textbf{73.7} & \textbf{36.2}& \textbf{46.1}& \textbf{54.7} & \underline{12.0}& \underline{28.8}& \textbf{9.8}& \textbf{37.3} & \textbf{18.3}\\

         \midrule
         \parbox[t]{2mm}{\multirow{1}{*}{\rotatebox[origin=c]{90}{FS}}}&Oracle&84.1 & 50.6& 60.7& 74.2 & 23.5& 56.9& 28.0& 54.0 & 29.1\\
        
         \bottomrule
    \end{tabular}
    }
    \vspace{0.1in}
    \captionsetup{font=small}
    \caption{\small \textbf{3D-FRONT $\rightarrow$ SUN RGB-D with per object category mAP@0.25 results.} 
    }
    \vspace{-3mm}
    \label{sunrgbd}
    
\end{table*}

\begin{figure}
\captionsetup[subfloat]{labelformat=empty}

\centering
\subfloat{
\centering
\includegraphics[width=0.24\linewidth]{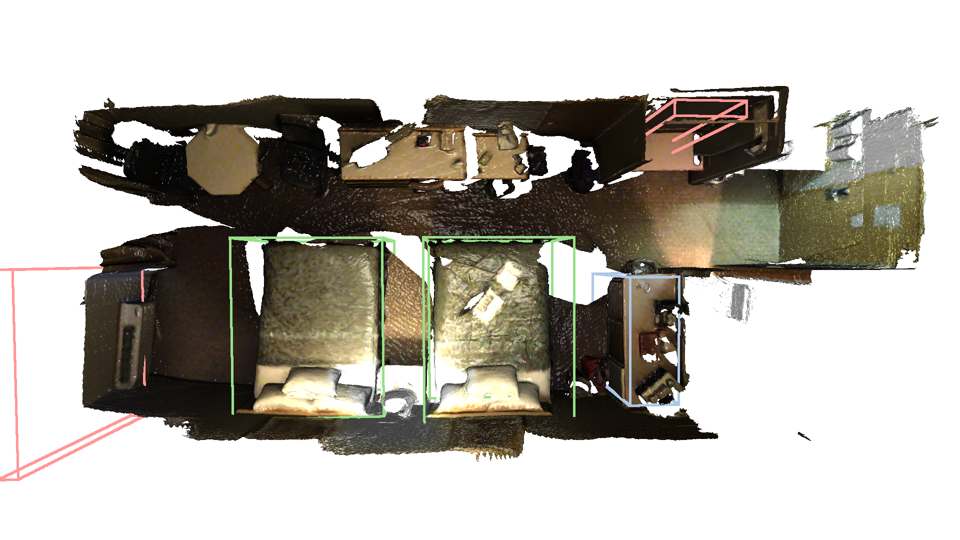}
}
\subfloat{
\centering

\includegraphics[width=0.24\linewidth]{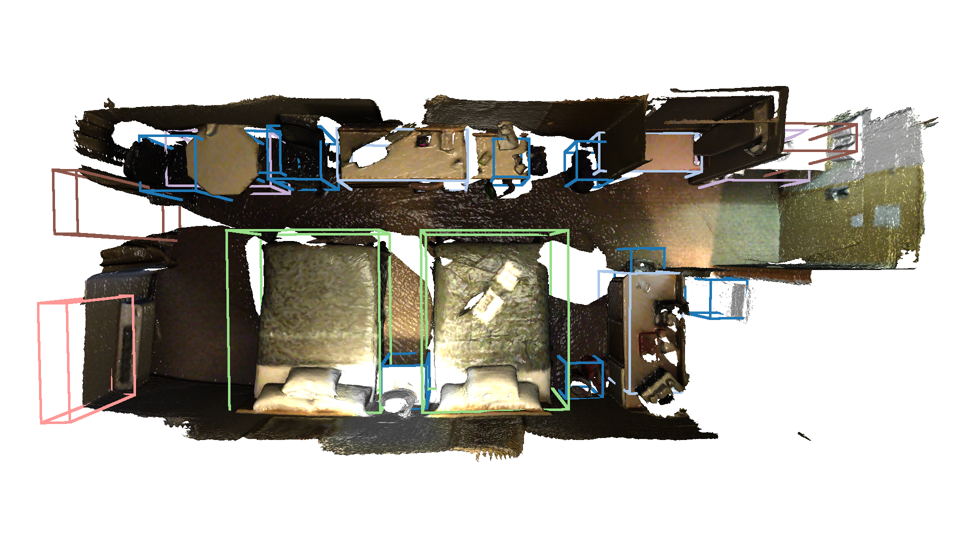}
}
\subfloat{
\centering
\includegraphics[width=0.24\linewidth]{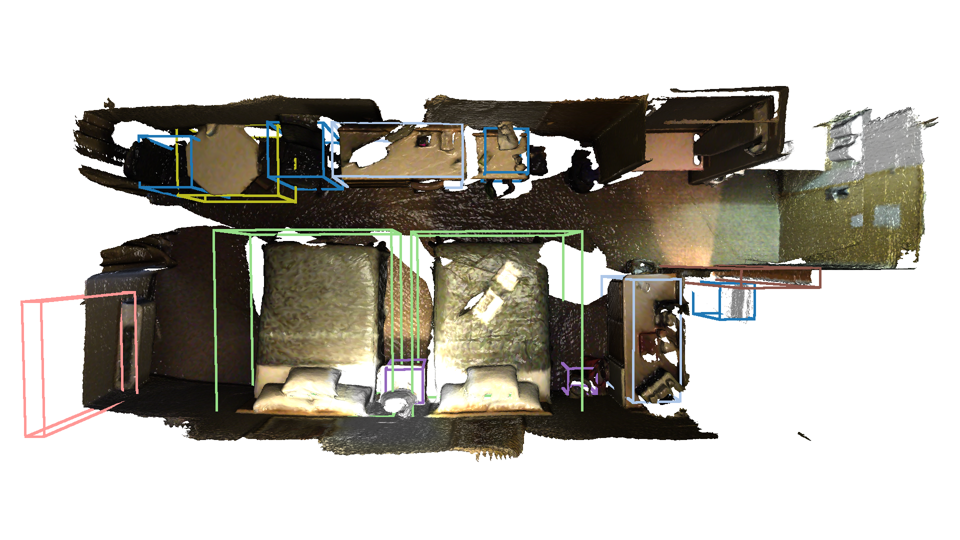}
}
\subfloat{
\centering

\includegraphics[width=0.24\linewidth]{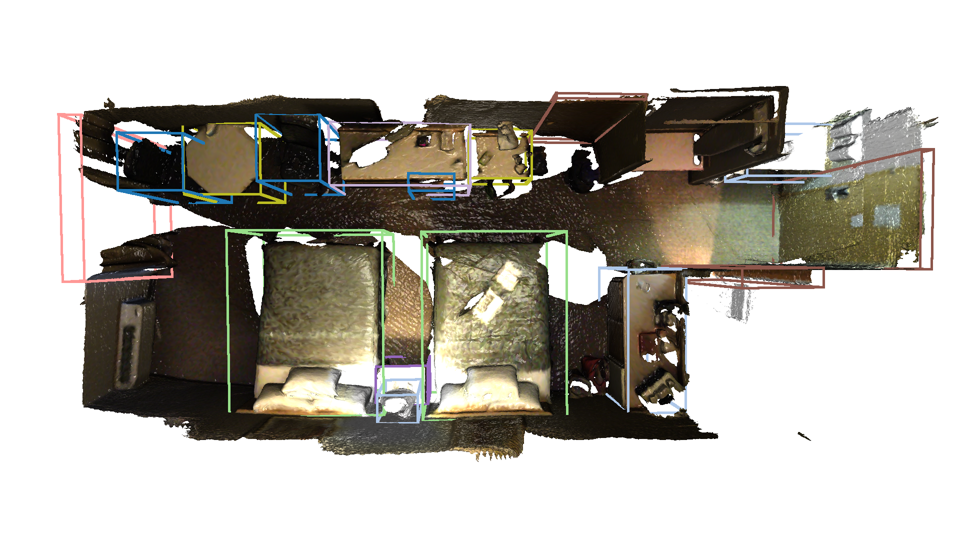}
}
\centerline{}
\subfloat[VSS Source Only]{
\centering
\includegraphics[width=0.24\linewidth]{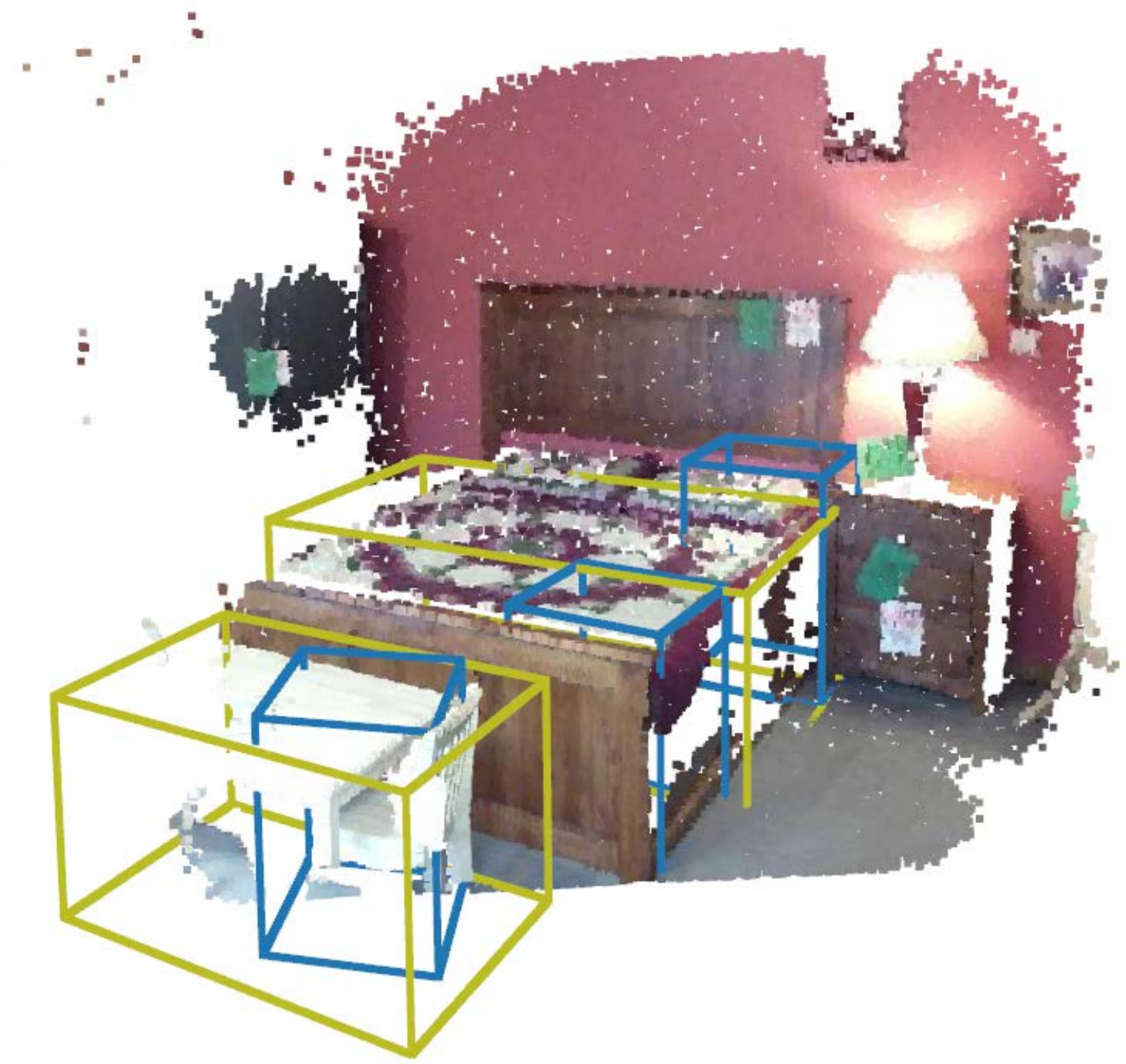}
}
\subfloat[ST3D]{
\centering

\includegraphics[width=0.24\linewidth]{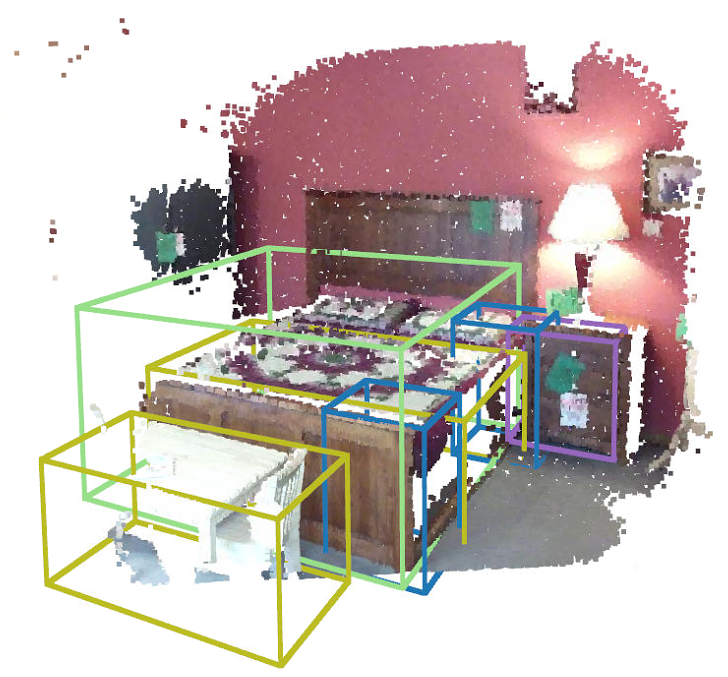}
}
\subfloat[Ours]{
\centering
\includegraphics[width=0.24\linewidth]{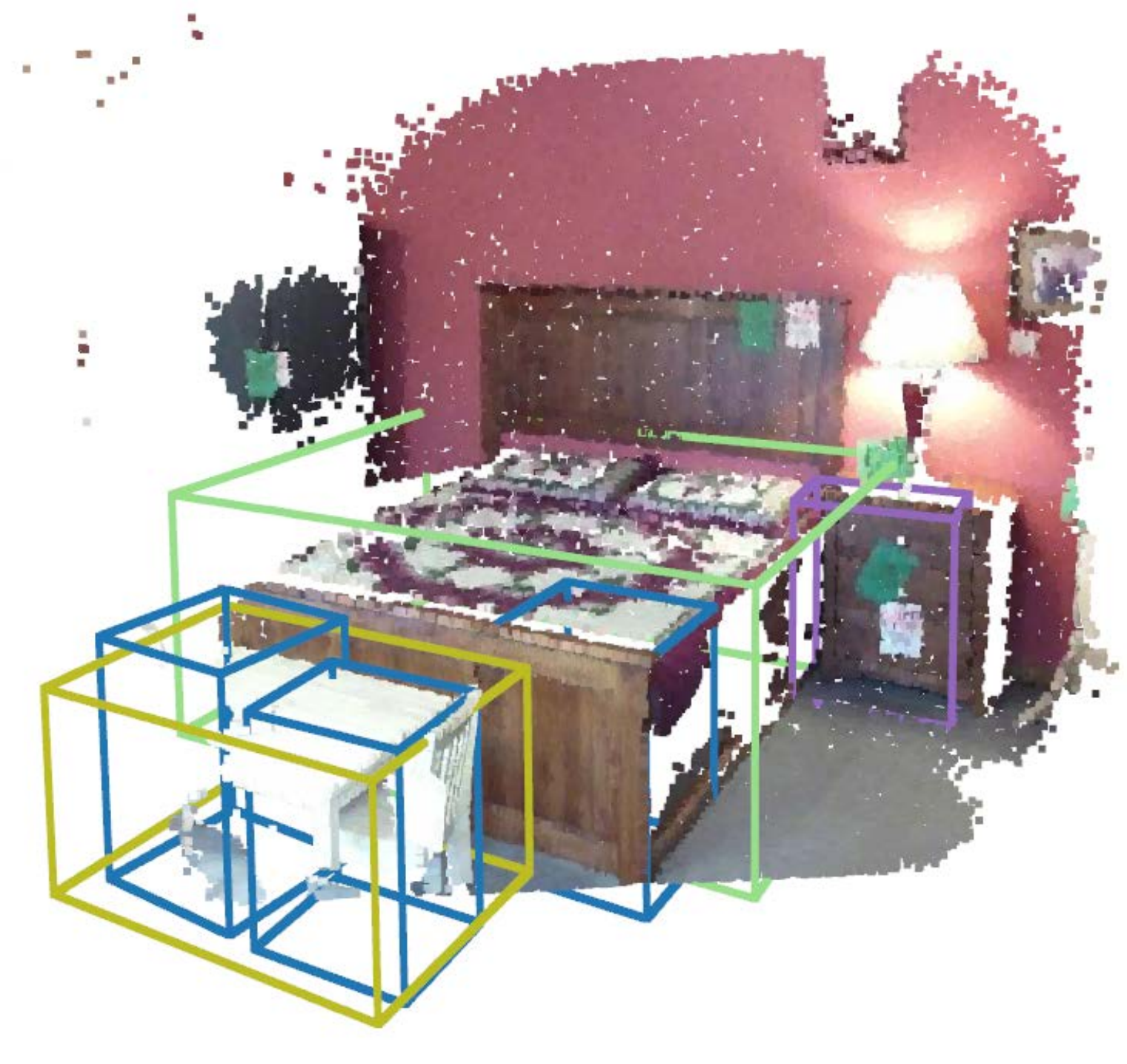}
}
\subfloat[Ground Truth]{
\centering

\includegraphics[width=0.24\linewidth]{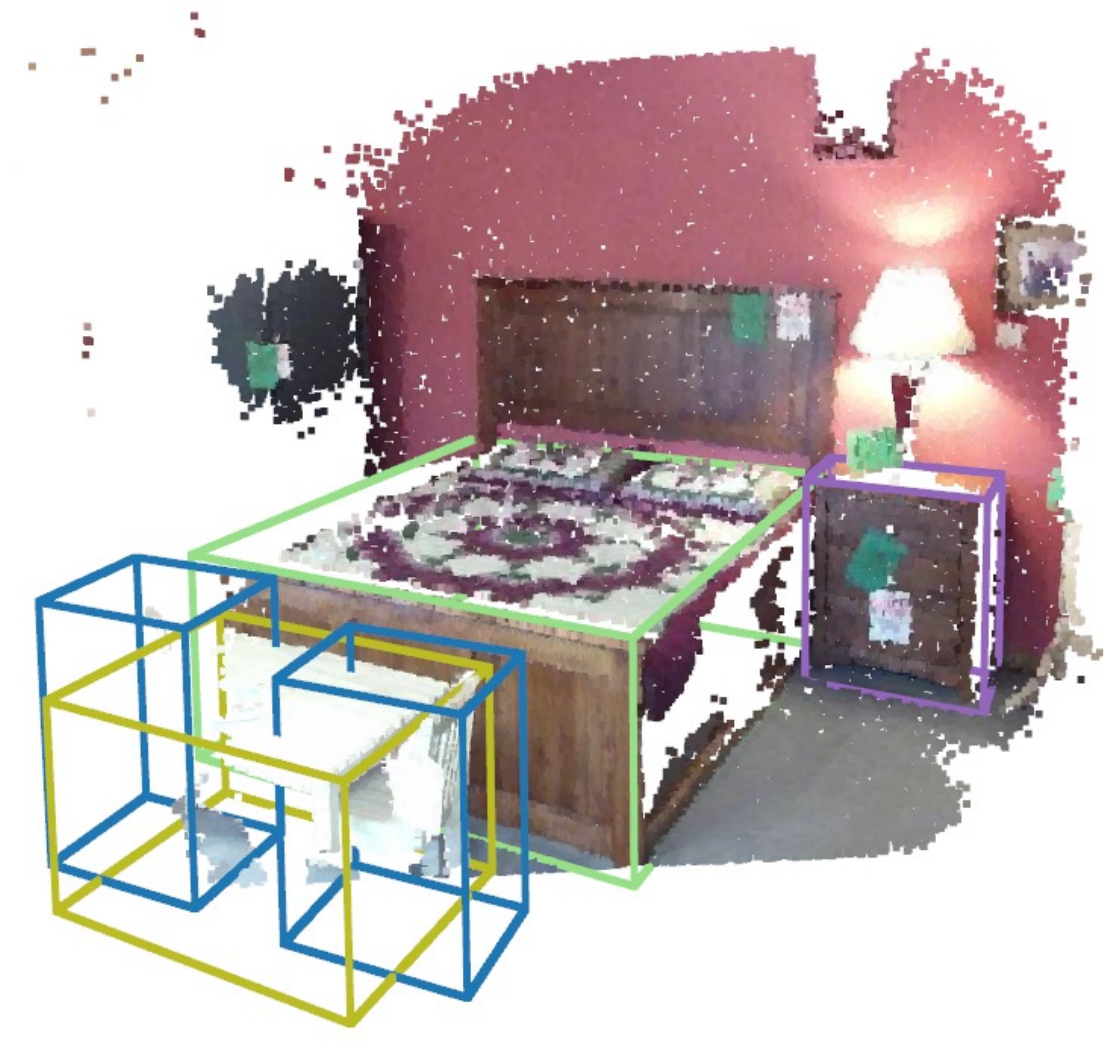}
}
\vspace{0.1in}
\caption{\small{\textbf{Qualitative results on ScanNetV2 (first row) and SUN RGB-D (second row).}}}
\vspace{-0.15in}
\label{qual}
\end{figure}

\noindent\textbf{ST3D \cite{st3d}}: We reproduce their adaptation results for indoor object detection using their consistency ensemble, where we only use box loss and semantic loss for self-training, similarly to our $\mathcal{L}_{cla}$. Their consistency ensemble can be regarded as a variant of our MPR module.

\noindent\textbf{Global-DAT \cite{adversarial}}: We average the seed point features and feed to a global discriminator to predict the domain, and perform min-max optimization similarly.
\subsection{Main Results}
\noindent\textbf{3D-FRONT $\rightarrow$ ScanNetV2 results.} 
As shown in Table \ref{scannet}, for data augmentation, OAA enhances mAP by $1\%\sim2\%$ upon integration with VSS \cite{doda}. In the UDA scenario, the direct application of naive MT \cite{mt} and Global-DAT \cite{adversarial} to the SR-UDA problem results in performance drop, and the ST3D only boosts mAP marginally by $1\sim 2\%$. In contrast, our OHDA \textit{w/o} PLR already yields substantial gains, and the introduction of PLR elevates the mAP by $\sim 4\%$, surpassing the VSS Source-Only baseline by $9.7\%$ mAP.

\vspace{2pt}
\noindent\textbf{3D-FRONT $\rightarrow$ SUN RGB-D results.} 
In Table \ref{sunrgbd}, under the 3D-FRONT $\rightarrow$ SUN RGB-D setting, our approach achieves an mAP increase of $9.1\%$ compared to the VSS-augmented source-only baseline, outperforming other UDA methods adapted from different contexts. 
Note that the improvements when adding OAA with VSS demonstrate its effectiveness in diversifying object layout distribution, which potentially enhances the learning of the object headings. 

\vspace{2pt}
\noindent\textbf{Qualitative results.} The qualitative comparisons are shown in Figure \ref{qual}. 
We demonstrate qualitative improvements in various object categories with different domain gaps, where we can predict more precise bounding boxes with higher recall.
\begin{figure}
\captionsetup{font=small}
\centering
\subfloat[\textit{w/o Aug}]{
\centering
\includegraphics[width=0.15\linewidth]{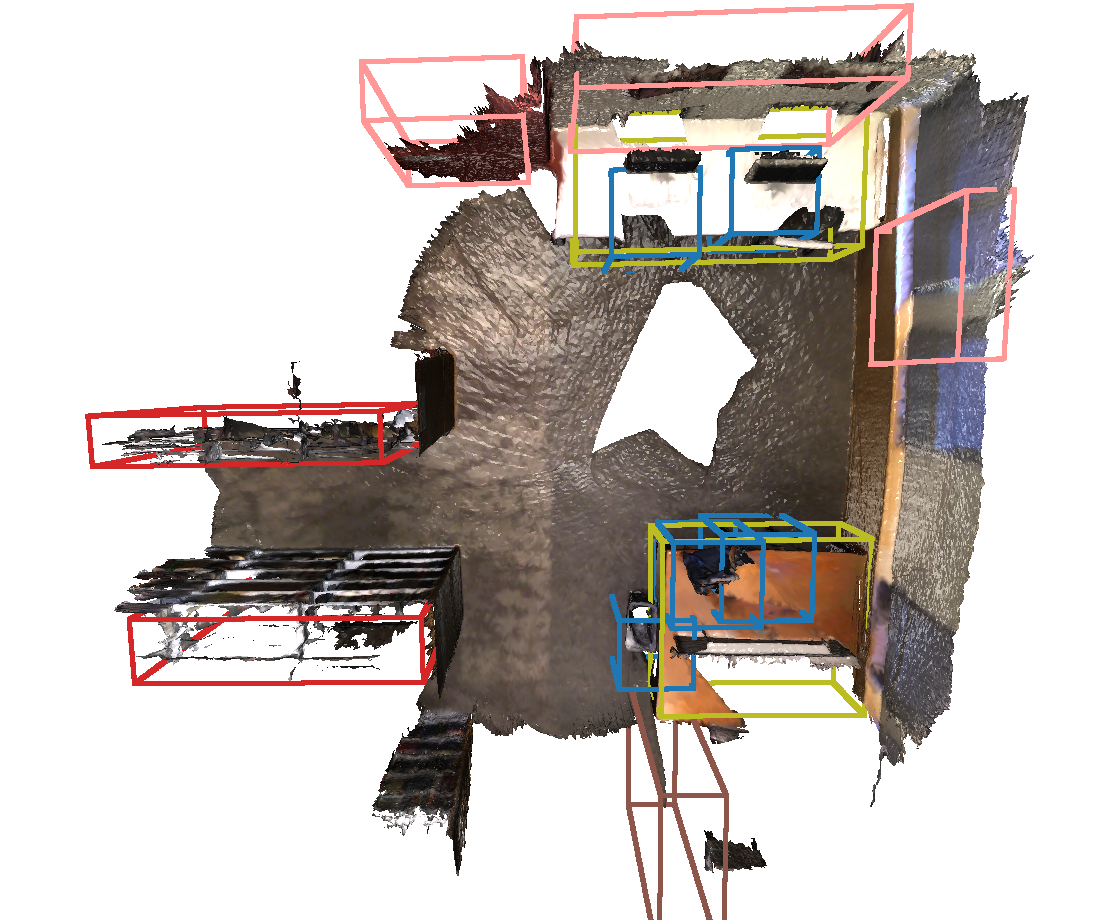}
}
\subfloat[\textit{w/o} CLA]{
\centering
\includegraphics[width=0.15\linewidth]{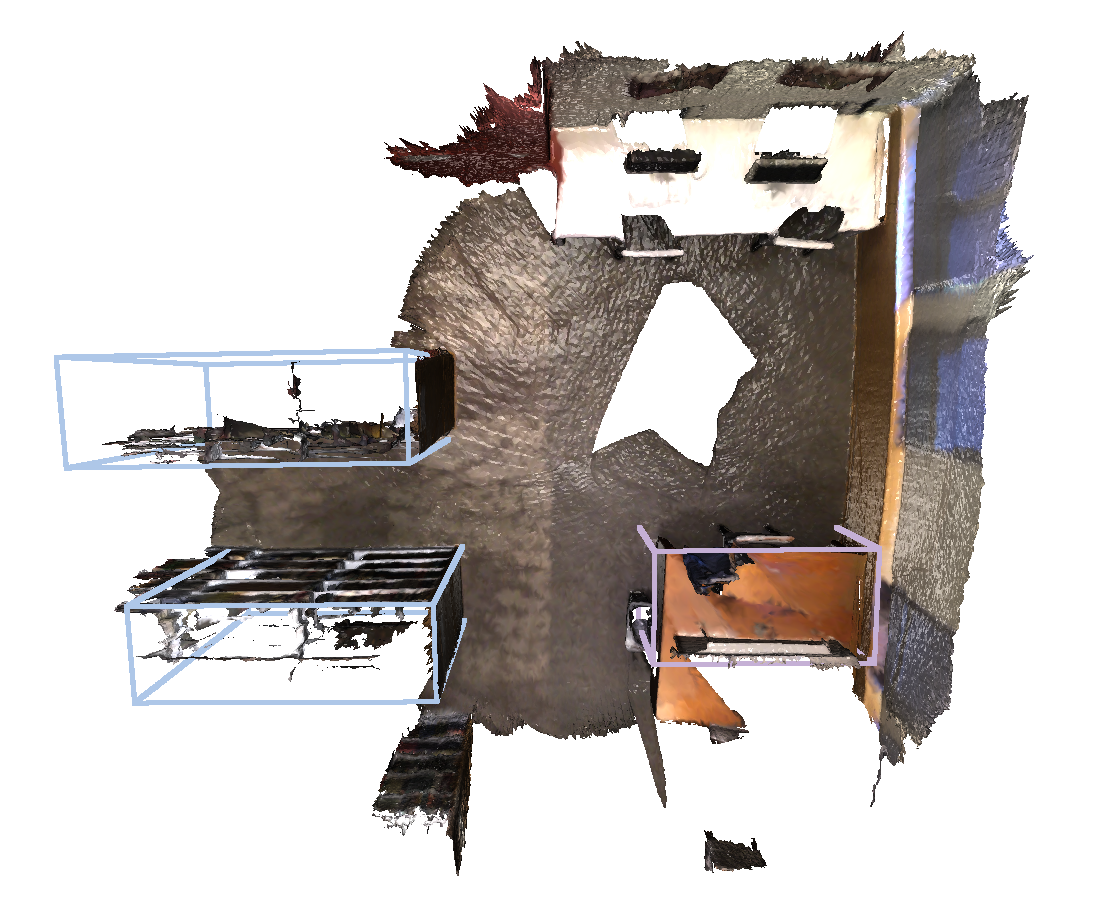}
}
\subfloat[\textit{w/o} HLA]{
\centering
\includegraphics[width=0.15\linewidth]{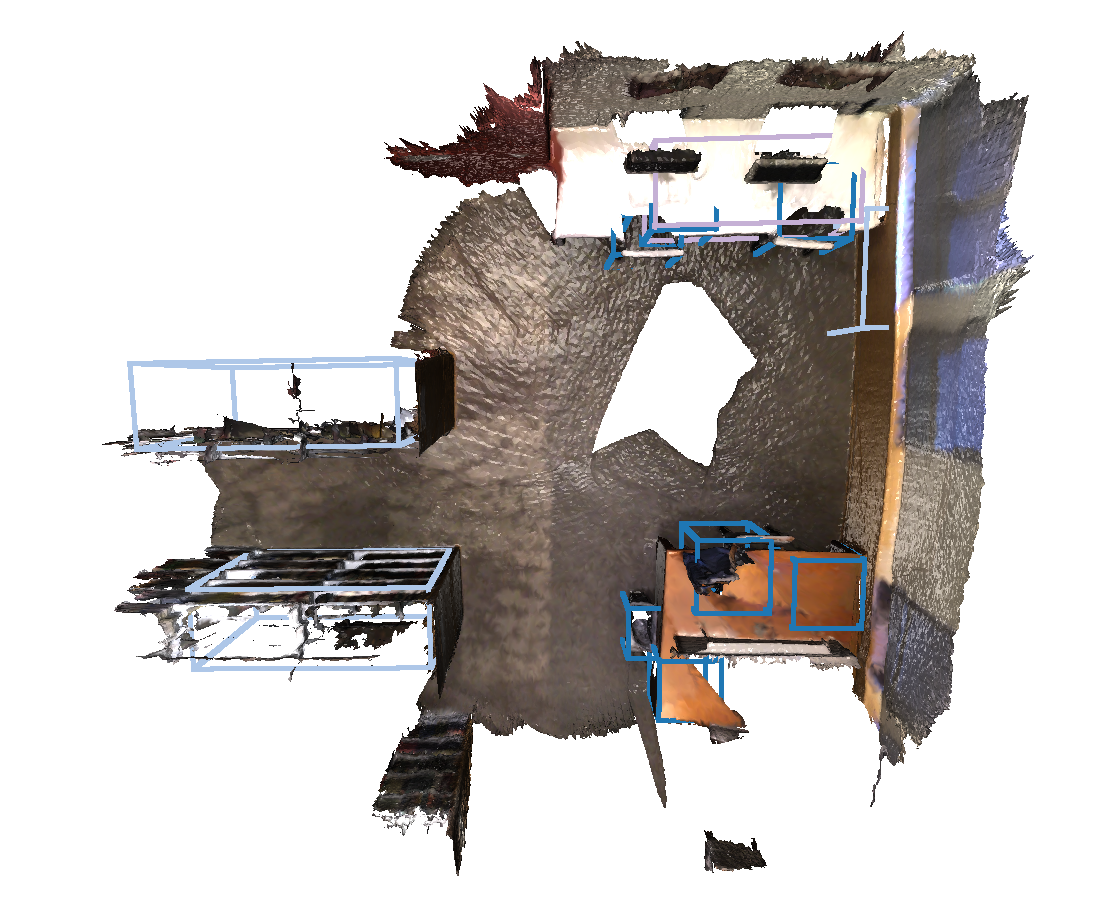}
}
\subfloat[\textit{w/o} PLR]{
\centering
\includegraphics[width=0.15\linewidth]{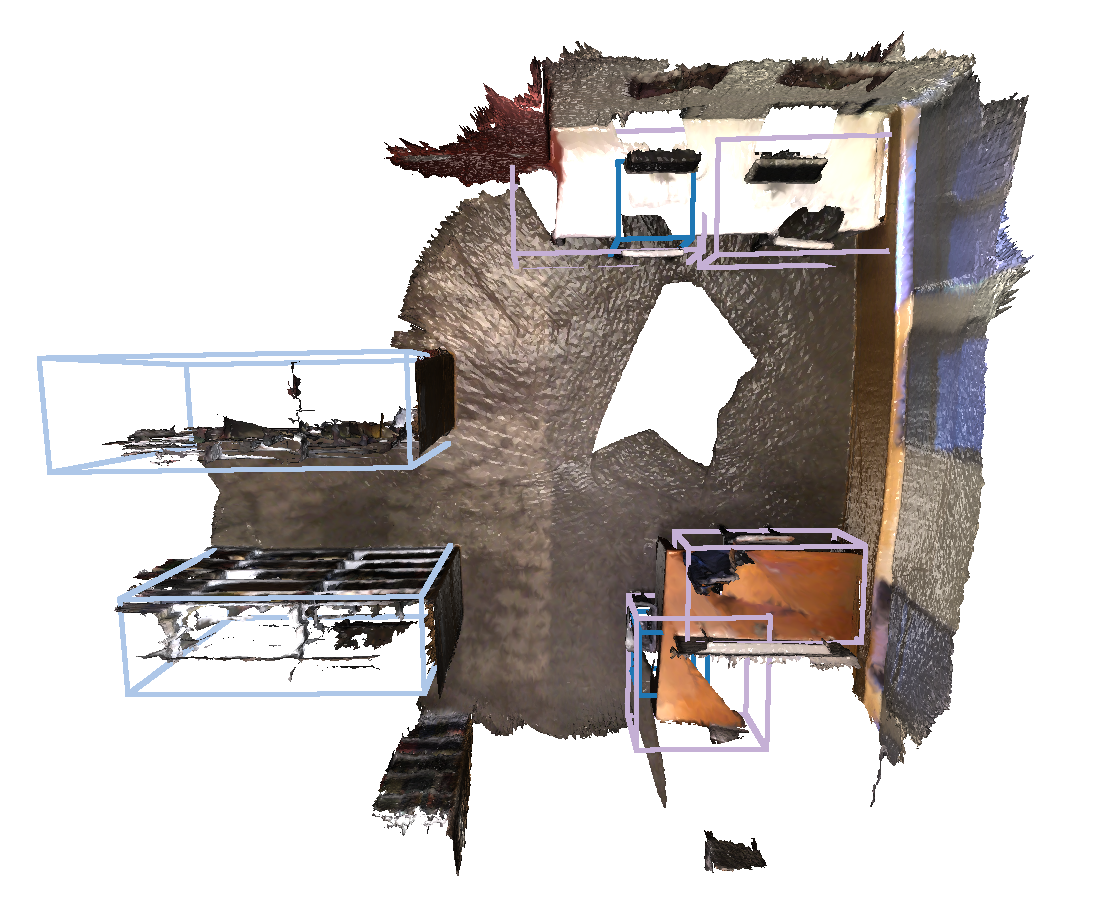}
}
\subfloat[Ours]{
\centering
\includegraphics[width=0.15\linewidth]{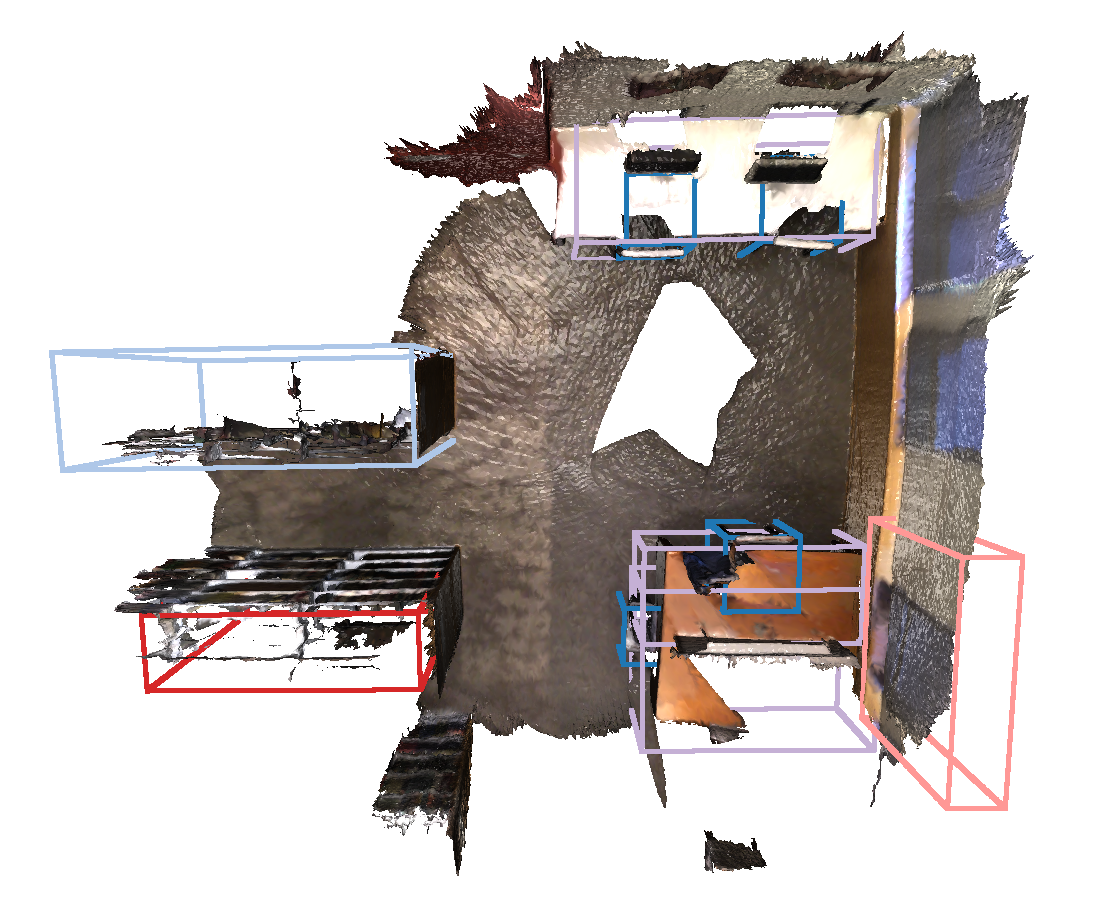}
}
\subfloat[Ground Truth]{
\centering
\includegraphics[width=0.15\linewidth]{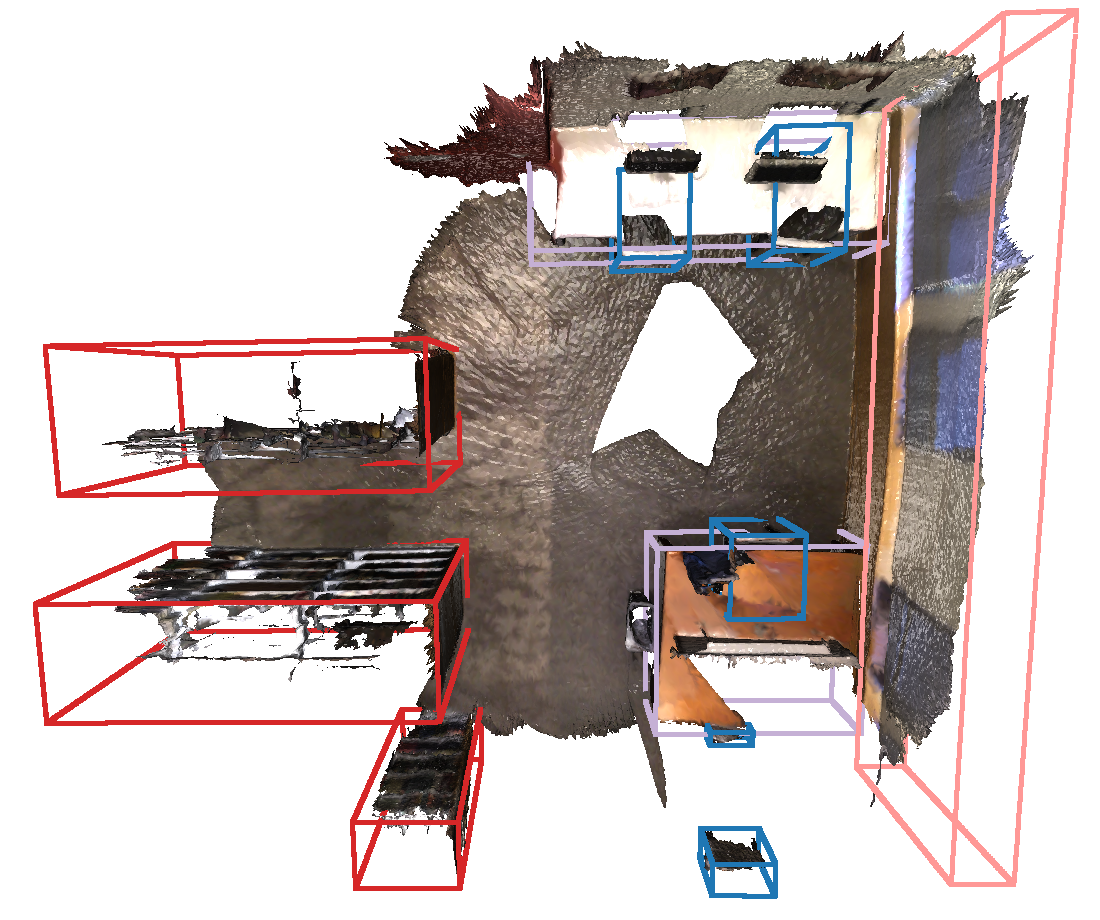}
}
\vspace{0.1in}
\caption{\small{\textbf{Qualitative results for ablation study.} Better viewed in color with zoom in.}}
\vspace{-0.15in}
\label{qual_ablation}
\end{figure}

\subsection{Ablation Study}

\noindent We conducted ablation study to assess the efficacy of the key components in our proposed model. This evaluation is detailed both quantitatively in Table \ref{tab:ablation} and qualitatively in Figure \ref{qual_ablation}. \begin{wraptable}{r}{7cm}
\footnotesize
    \captionsetup{font=small}
\setlength{\tabcolsep}{1mm}
    {
    \begin{tabular}{cccc|cc|cc}
        \toprule
         \multirow{2}{*}{\textit{Aug.}} & \multirow{2}{*}{CLA} & \multirow{2}{*}{HLA} & \multirow{2}{*}{PLR} &\multicolumn{2}{c|}{ScanNetV2} & \multicolumn{2}{c}{SUN RGB-D} \\
         & & &&mAP$_{25}$ & mAP$_{50}$&mAP$_{25}$ & mAP$_{50}$ \\
         \midrule
         \ding{56} &\ding{51} & \ding{51} & \ding{51} & 37.0 & 22.2 & 29.3 & 13.5  \\
         \ding{51} &\ding{56} & \ding{51} & \ding{51}$^*$ & 35.6 & 22.0 & 31.6 & 13.4  \\
         \ding{51} &\ding{51} & \ding{56} & \ding{56} &  35.9 & 21.0 & 33.1 & 14.7   \\
         \ding{51} &\ding{51} & \ding{56} & \ding{51} &  \underline{39.3} & \underline{24.1} & \underline{35.5} & \underline{16.8}   \\
         \ding{51} &\ding{51} & \ding{51} & \ding{56} &  38.5 & 23.9 &  \underline{35.5} & 15.6   \\
         \ding{51} &\ding{51} & \ding{51} & \ding{51} & \textbf{42.9}  & \textbf{27.6} & \textbf{37.3} & \textbf{18.3} \\
         \bottomrule
    \end{tabular}}
    \vspace{-0.0in}
    \caption{\small{\textbf{Ablation study.} $^*$ means the PLR components are invalid without CLA.}}
    \label{tab:ablation}
\end{wraptable}The results substantiate the necessity of each component in our proposed methodology. Furthermore, an integration of the results from Table \ref{scannet} and Table \ref{sunrgbd} reveals a notable insight: while domain adversarial training and naive pseudo labeling might lead to only marginal improvements or even a decline in performance when implemented separately, their synergistic application markedly boosts overall performance. This compellingly underscores the merits of our proposed OHDA framework. Additional ablation studies are available in the supplementary material.

\section{Conclusion}

In this paper we present OHDA framework, a pioneering solution to the SR-UDA challenge in indoor 3D object detection. 
Our key contribution is the integration of proposal-level domain adversarial training and pseudo labeling, which simultaneously align cross-domain proposal feature space and perform effective joint-training under complicated domain shifts. 
Extensive experiments underline the impressive efficacy of our approach in tackling the SR-UDA problem, where we achieve $9\%$\,$\sim$\,$ 10\%$ consistent mAP gains over Source-Only baselines and consistently surpass existing UDA methods adapted from outdoor scenario. We hope that our work could inspire further research in this direction.

\section{Acknowledgement}
This work is supported by the Agency for Science, Technology and Research (A*STAR) under its MTC Programmatic Funds (Grant No. M23L7b0021).

\bibliography{egbib}
\end{document}